\documentclass[letterpaper, 10 pt, conference]{ieeeconf}  

\IEEEoverridecommandlockouts                              

\usepackage{graphicx}
\usepackage{amsmath} 
\usepackage{amssymb}  
\usepackage{booktabs}
\usepackage{float}

\title{\LARGE \bf
LIN-SLAM: Lightweight LiDAR--Inertial Neural SLAM for Real-Time Onboard Navigation
}

\author{
Shiqi Che, Yijie Liu, Shuo Wang, and Wenxin Liu
\thanks{
This work was supported in part by the National Natural Science Foundation
of China under Grant T2596034, and by the Fundamental Research Funds for
the Central Universities.
}
\thanks{
Shiqi Che, Yijie Liu, Shuo Wang, and Wenxin Liu are with the
School of Automation Science and Electrical Engineering, Beihang University,
Beijing 100191, China
(e-mail: sqche@buaa.edu.cn; yijiel@buaa.edu.cn; sWangsuki@buaa.edu.cn;
wenxinl@buaa.edu.cn).
Corresponding author: Wenxin Liu
}
}

\begin{document}

\maketitle
\thispagestyle{empty}
\pagestyle{empty}

\begin{abstract}

Geometric light detection and ranging (LiDAR)--inertial simultaneous
localization and mapping (SLAM) provides efficient state estimation but lacks the continuous map representation and global
refinement capabilities of neural implicit SLAM, while neural implicit methods remain computationally expensive. We propose Lightweight LiDAR--Inertial Neural SLAM (LIN-SLAM), a hybrid method that couples
high-rate geometric LiDAR--inertial estimation with neural implicit
tracking, continual mapping, and global refinement for accurate real-time
onboard SLAM. Experiments on twelve representative sequences from three datasets spanning 32-, 64-, and 128-channel LiDAR configurations show that LIN-SLAM achieves the lowest overall mean absolute trajectory error (ATE) among the evaluated state-of-the-art geometric and neural implicit methods, while substantially reducing runtime compared with existing neural implicit SLAM methods. On a Unitree Go2 equipped with an NVIDIA Jetson Orin NX and a 16-channel LiDAR, LIN-SLAM sustains real-time localization and neural mapping over a 1.47-km indoor trajectory through challenging repetitive corridor environments, with an endpoint closure error of only 0.141~m. 
To the best of our knowledge, LIN-SLAM is the first LiDAR--inertial SLAM system with a neural implicit backend to achieve real-time online operation on a mobile robotic platform while maintaining state-of-the-art localization accuracy. The code will be released upon acceptance.

\end{abstract}

\section{INTRODUCTION}

Mobile robots require real-time state estimation and globally
consistent mapping within limited onboard computational and memory
budgets.
Geometric LiDAR--inertial odometry (LIO)
provides efficient local
motion estimation through point registration and inertial
fusion~\cite{xu2022fastlio2,qu2022llol,bai2022fasterlio}. However, its local motion
constraints provide limited correction of accumulated drift and
inconsistencies across revisited locations, motivating map-level
refinement.
Neural implicit SLAM integrates
continuous surface representations with
map-based tracking, continual reconstruction, and global map
refinement~\cite{pan2024pinslam,jiang2025clidslam}. However, these capabilities
require repeated signed distance field (SDF) queries, gradient
evaluation, and online optimization, introducing substantial computational
and memory overhead. Sustaining this workload alongside responsive
state estimation is particularly challenging on resource-constrained
onboard platforms. Achieving the accuracy and map consistency of
neural implicit SLAM while maintaining low computational cost within
onboard resource constraints therefore remains an open challenge.

We present LIN-SLAM, a hybrid framework that combines the best of both worlds: the computational efficiency of traditional geometric methods with the map consistency and high accuracy of neural implicit approaches.  
LIN-SLAM consists of three complementary components. 
The geometric frontend provides responsive motion estimates and
posterior-consistent scans. The keyframe interface transfers geometry,
motion, and uncertainty to the neural implicit backend. The neural implicit backend refines
poses, continually learns a continuous map, and maintains global
consistency, while asynchronous operation supports real-time
onboard execution.

The geometric frontend further improves neural implicit tracking and
mapping by providing posterior-conditioned geometry, relative motion,
and uncertainty. These outputs provide better-aligned observations and
reliable initialization for the neural implicit backend, reducing
trajectory estimation error and improving map reconstruction quality.
In return, the accepted global map revisions correct the high-rate frontend tracking through the map-to-odometry transform without interrupting the local estimator.
We evaluate our hybrid framework on twelve public sequences from Newer College Stereo-Cam, Newer College Multi-Cam, and M2DGR, spanning 32-, 64-, and 128-channel LiDAR configurations. The results show that LIN-SLAM achieves the lowest overall mean ATE among all evaluated geometric and neural implicit methods.

We further optimize our hybrid approach for computational efficiency, enabling real-time performance on resource-constrained edge computing platforms.
We successfully deploy LIN-SLAM on a Unitree Go2 equipped with an NVIDIA Jetson Orin NX and a Hesai 16-channel LiDAR, demonstrating real-time localization and neural mapping over kilometer-scale trajectories in challenging indoor environments with low endpoint closure error. In contrast, the evaluated geometric baselines accumulate substantial drift or diverge on the challenging onboard trajectories, while the evaluated neural implicit baselines cannot execute the complete SLAM pipeline under the same onboard resource constraints. To the best of our knowledge, LIN-SLAM is the first LiDAR--inertial SLAM system with a neural implicit backend to achieve real-time online operation on a resource-constrained mobile robotic
platform.

Our contributions are threefold:

\begin{itemize}

\item \textbf{A hybrid framework for LiDAR--inertial neural implicit SLAM.}
We integrate high-rate geometric LiDAR--inertial estimation with neural
tracking, mapping, and global refinement through bidirectional
frontend--backend information exchange. LIN-SLAM achieves the lowest
overall mean ATE across the evaluated public sequences and endpoint closure errors below 0.15~m over two real-time onboard trajectories.

\item \textbf{Lightweight optimization for real-time onboard deployment.}
We decouple geometric estimation and neural implicit processing into asynchronous threads to preserve estimation accuracy while maintaining computational efficiency. Selective neural-backend processing together with an optimized implementation in C++ and CUDA substantially reduces complete-system processing cost, enabling real-time onboard operation.

\item \textbf{An open-source implementation of LIN-SLAM.}
We will release LIN-SLAM as open-source software to facilitate reproducibility, real-world deployment, and further research on lightweight neural implicit SLAM.

\end{itemize}

\section{RELATED WORK}

LiDAR--inertial odometry has progressively moved toward tightly coupled estimation with higher update bandwidth 
and more accurate motion compensation. FAST-LIO~\cite{xu2021fastlio} formulates LiDAR--inertial fusion within an iterated Kalman-filter 
framework, where geometric LiDAR observations update the inertially propagated state, while FAST-LIO2~\cite{xu2022fastlio2} further removes 
explicit geometric feature extraction and directly registers raw LiDAR measurements to an incrementally maintained 
map, providing an efficient tightly coupled formulation applicable to different LiDAR scanning patterns. 
Point-LIO pushes the estimation bandwidth further through point-wise state updates at the actual 
sampling time of each LiDAR measurement, thereby avoiding accumulation and providing high-frequency 
odometry under aggressive motion~\cite{he2023pointlio}. Together, these methods establish tightly 
coupled geometric LiDAR--inertial filtering as an effective basis for high-bandwidth state estimation.

Another line of work focuses on the temporal structure and motion distortion within LiDAR scans. 
DLIO constructs analytical continuous-time trajectories for point-wise motion correction and combines 
them with direct scan-to-map registration, targeting accurate yet lightweight odometry under aggressive 
motion~\cite{chen2023dlio}. SLICT similarly models the LiDAR--inertial trajectory in continuous time and
employs surfel-based registration to exploit geometric constraints throughout the scan acquisition 
interval~\cite{nguyen2023slict}. For spinning LiDARs, LLOL reduces sensing latency by processing partial 
sweeps as a continuous stream instead of waiting for complete physical revolutions~\cite{qu2022llol}. 
Its local trajectory is repeatedly propagated during iterative registration, enabling pose updates at 
rates substantially higher than the LiDAR revolution frequency. These approaches demonstrate 
the benefits of high-rate estimation, continuous-time motion modeling, and partial-sweep processing 
for reducing the latency and motion distortion inherent in frame-based LiDAR estimation.

More recently, neural implicit representations have been introduced into LiDAR SLAM to provide continuous geometric modeling while improving odometry and map consistency.
SHINE-Mapping~\cite{zhong2023shine} demonstrates large-scale incremental reconstruction with sparse implicit features, while NeRF-LOAM~\cite{deng2023nerfloam} extends neural SDFs to joint LiDAR odometry and mapping.
PIN-SLAM further develops a unified point-based implicit representation for correspondence-free tracking, continual mapping, loop detection, pose-graph optimization, and map deformation~\cite{pan2024pinslam}. Its elastic neural map enables globally consistent dense reconstruction and accurate tracking. NDF-SLAM~\cite{yang2025ndfslam} and S$^2$KAN-SLAM~\cite{wang2025s2kan} use neural distance fields and SDF submaps, respectively. However, these methods primarily operate on LiDAR measurements alone, and typically rely on discrete GPU computation, posing challenges for real-time deployment on resource-constrained onboard platforms.

Neural implicit SLAM has also been extended to incorporate inertial measurements for more robust state estimation. CLID-SLAM tightly couples neural SDF representations with LiDAR--inertial estimation by introducing neural-SDF observations into an iterated error-state Kalman filter (IESKF) and employing region-specific supervision for geometric modeling~\cite{jiang2025clidslam}. While such tight coupling improves the integration of neural mapping and motion estimation, neural-field evaluation is directly involved in the measurement update, introducing additional computational overhead into the estimation pipeline. Moreover, its evaluation focuses on workstation-class discrete-GPU platforms, leaving real-time deployment on resource-constrained onboard hardware largely unexplored.

\begin{figure*}[t]
   \centering
   \includegraphics[width=\textwidth]{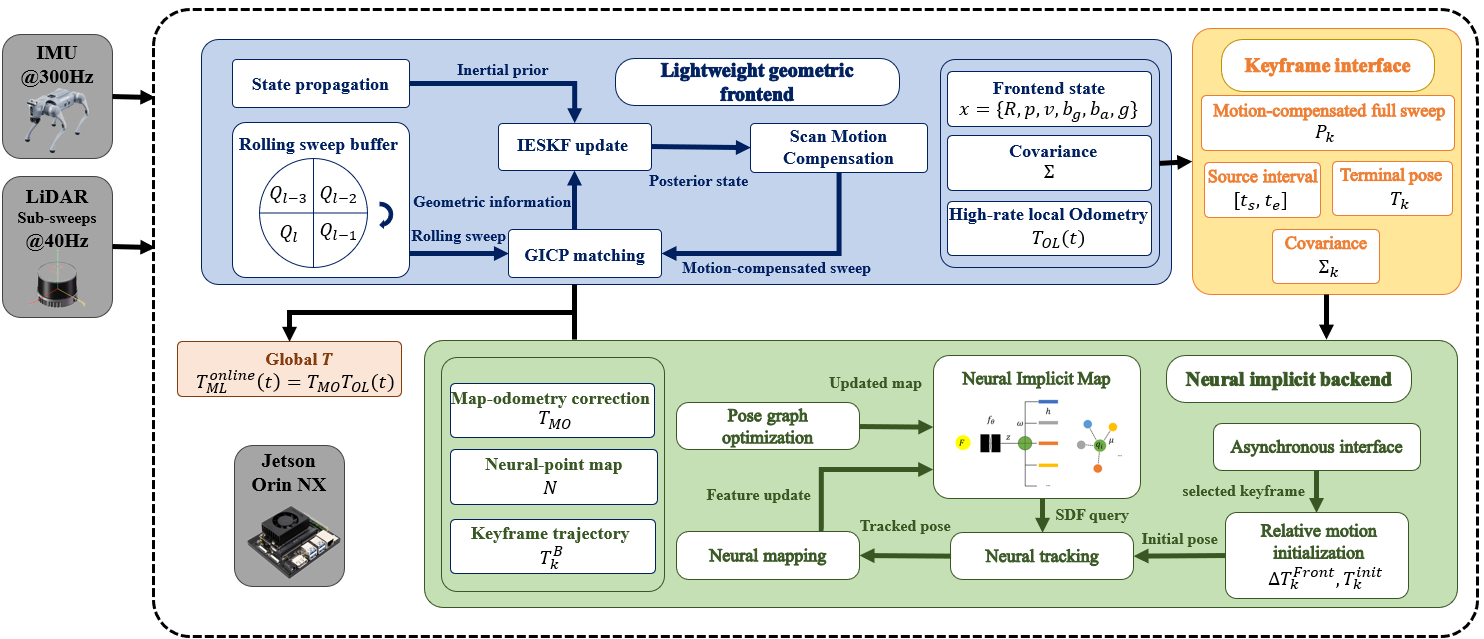}
   \caption{Overview of the lightweight LIN-SLAM pipeline. A geometric generalized iterative closest point (GICP)--IESKF frontend processes LiDAR sub-sweeps and IMU measurements to provide local poses and motion-compensated scans. Full-sweep keyframe candidates associate geometry with source intervals, terminal poses, and covariances for asynchronous transfer. The neural implicit backend uses optimized C++/CUDA computation to reduce processing overhead, fixed-capacity replay storage and local neural windows to control stored data, and preallocated workspaces to reduce repeated allocation. Frontend relative motion initializes neural tracking; continual mapping and loop-based global optimization update the neural map. Backend corrections provide globally corrected poses at the frontend rate.}
   \label{fig:system}
\end{figure*}

\section{LIN-SLAM}
\label{sec:method}

We propose LIN-SLAM for real-time onboard navigation by coupling
high-rate geometric LiDAR--inertial estimation with neural implicit
mapping (Fig.~\ref{fig:system}). An asynchronous keyframe interface
transfers motion-compensated sweeps, poses, and covariances to the
backend for tracking, continual mapping, and global optimization
without blocking the frontend.

We use the LiDAR frame $L$, inertial measurement unit (IMU)/body frame $I$, coordinate frame $P$
of the local depth panorama, fixed odometry frame $O$, and global map
frame $M$. The depth panorama stores LiDAR depth observations by
viewing direction.
The IMU has orientation $\mathbf R_{PI}$, position $\mathbf p_{PI}$,
and velocity $\mathbf v_P$ in $P$.
We use $\mathbf T_{AB}\in SE(3)$ to denote the rigid transformation
from frame $B$ to frame $A$, with
$\mathbf p^A=\mathbf T_{AB}\odot\mathbf p^B
=\mathbf R_{AB}\mathbf p^B+\mathbf t_{AB}$. Therefore, the frontend transform chain gives
\begin{equation}
\mathbf T_{OL}(t)=
\mathbf T_{OP}(t)\mathbf T_{PI}(t)\mathbf T_{IL} ,
\label{eq:frame_chain}
\end{equation}
where $\mathbf T_{IL}$ is the fixed LiDAR-to-IMU extrinsic transform.
The frontend resets the origin of the gravity-aligned panorama frame
$P$ without changing its axes; $O$ remains fixed.
The transform $\mathbf T_{MO}$ represents corrections after global
backend optimization.


\subsection{Lightweight Geometric Frontend}
\label{sec:lightweight_frontend}

\subsubsection{State Definition and Propagation}

The IESKF retains an 18-dimensional error state, with
$\mathbf g_P$ denoting gravity in $P$:
\begin{equation}
\begin{aligned}
\mathbf x
&=
(\mathbf R_{PI},\mathbf p_{PI},\mathbf v_P,
\mathbf b_g,\mathbf b_a,\mathbf g_P),\\
\delta\mathbf x
&=
[\delta\boldsymbol\theta^T,\delta\mathbf p^T,
\delta\mathbf v^T,\delta\mathbf b_g^T,
\delta\mathbf b_a^T,\delta\mathbf g^T]^T .
\end{aligned}
\label{eq:state}
\end{equation}
Rotation updates use right multiplication; other state updates are
additive. With the IMU stationary and level, the frontend initializes
the gyroscope and accelerometer biases $\mathbf b_g,\mathbf b_a$ and
fixes gravity along the nominal $+Z$ direction of $P$. It retains the
gravity state and covariance blocks but disables LiDAR-induced
gravity updates.
Measured angular rate $\boldsymbol\omega_m$ and specific force
$\mathbf a_m$ give bias-corrected inputs
$\boldsymbol\omega=\boldsymbol\omega_m-\mathbf b_g$ and
$\mathbf f=\mathbf a_m-\mathbf b_a$.
IMU integration propagates the nominal state and covariance,
yielding the inertial prior $(\mathbf x^-,\boldsymbol\Sigma_-)$
~\cite{xu2021fastlio,xu2022fastlio2}.

\subsubsection{Rolling-Sweep Buffer}

Following LLOL~\cite{qu2022llol}, the active sweep
$\mathcal S_\ell=\bigcup_{i=0}^{3}\mathcal Q_{\ell-i}$ combines the
$\ell$-th quarter-sweep $\mathcal Q_\ell$ with the preceding three.
Before inserting each new quarter-sweep, the frontend fuses the oldest
sector into the local depth panorama map and replaces it. The other
sectors remain, providing approximately full-field-of-view scans and
pose updates above the revolution rate. The panorama accumulates
historical observations over the surrounding spherical field of view,
beyond a single revolution.

\subsubsection{GICP Matching}

GICP~\cite{segal2009gicp} uses the active rolling sweep
$\mathcal S_\ell$ as the source and the local depth panorama as
the target; the latter is separate from the backend neural map.
For each matched point pair $i$, $\mathbf r_i$, $\mathbf J_i^{\rm G}$,
$\boldsymbol\Omega_i$, and $\rho_i$ are the residual, pose Jacobian,
information matrix, and robust weight, respectively. With $\mathbf W_i=s_L\rho_i\boldsymbol\Omega_i$, the approximate
Hessian $\mathbf H_L$ and gradient $\mathbf g_L$ of the GICP
objective are
\begin{equation}
\mathbf H_L
=
\sum_i(\mathbf J_i^{\rm G})^T\mathbf W_i\mathbf J_i^{\rm G},
\qquad
\mathbf g_L
=
\sum_i(\mathbf J_i^{\rm G})^T\mathbf W_i\mathbf r_i ,
\label{eq:gicp_information}
\end{equation}
where $s_L$ controls the LiDAR information scale.

\subsubsection{IESKF Update}

The IESKF combines GICP constraints with the inertial prior to correct
the frontend state. The matrix $\mathbf E\in\mathbb R^{6\times18}$ selects attitude and
position errors from the error state. At iteration $j$, $\mathbf e_j$
is the accumulated error from the propagated prior to the current
linearization point. The matrix
$\mathbf\Gamma=\operatorname{diag}
(\mathbf I_3,\mathbf I_3,\mathbf I_3,
\mathbf 0_3,\mathbf 0_3,\mathbf 0_3)$
suppresses LiDAR-induced bias and gravity increments. The IESKF updates
the state and covariance as
\begin{equation}
\begin{aligned}
\mathbf A_j
&=\boldsymbol\Sigma_-^{-1}
+\mathbf E^T\mathbf H_L\mathbf E,\\
\mathbf A_j\delta\mathbf x_j
&=-\left(
\mathbf E^T\mathbf g_L+
\boldsymbol\Sigma_-^{-1}\mathbf e_j
\right),\\
\mathbf x_{j+1}
&=\mathbf x_j\boxplus
(\mathbf\Gamma\delta\mathbf x_j),\\
\mathbf C
&=\boldsymbol\Sigma_--\mathbf A_J^{-1},\\
\boldsymbol\Sigma_+
&=\mathbf A_J^{-1}
+(\mathbf I-\mathbf\Gamma)\mathbf C
(\mathbf I-\mathbf\Gamma)^T ,
\end{aligned}
\label{eq:ieskf_update}
\end{equation}
where $\boxplus$ applies the error-state increment, $J$ is the final
accepted iteration, and $\mathbf I$ is the $18\times18$ identity matrix.
GICP constrains attitude and position; pose--velocity cross-covariance
also allows velocity correction. The partial-gain covariance update
accounts for the selective correction through $\mathbf\Gamma$ and
provides pose covariance for the neural tracking optimization schedule.
The frontend updates correspondences and residual linearizations after
each correction until convergence or the iteration limit, rejecting
updates with insufficient valid GICP constraints. The accepted posterior
$(\mathbf x^+,\boldsymbol\Sigma_+)$ supports motion compensation and
keyframe transfer.

\subsubsection{Scan Motion Compensation}

After each accepted IESKF correction, the frontend reconstructs the
trajectory over $[t_s,t_e]$ to motion-compensate the rolling sweep before
the next GICP matching step. This interval covers the full sweep once
scan history, IMU measurements, and valid trajectory nodes support it;
during startup, the frontend uses only the supported portion.
For consecutive IMU times $t_0,t_1$, define $\Delta t=t_1-t_0$,
$\boldsymbol\alpha_\omega=
(\boldsymbol\omega_1-\boldsymbol\omega_0)/\Delta t$, and
$\mathbf j=(\mathbf a_1-\mathbf a_0)/\Delta t$,
using bias-corrected endpoint angular rates $\boldsymbol\omega_{0,1}$
and translational accelerations $\mathbf a_{0,1}$ in $P$.
With $\mathbf R_0=\mathbf R_{PI}(t_0)$,
$\mathbf p_0=\mathbf p_{PI}(t_0)$, and
$\mathbf v_0=\mathbf v_P(t_0)$, the trajectory for
$\tau\in[0,\Delta t]$ is
\begin{equation}
\begin{aligned}
\Delta\boldsymbol\theta(\tau)
&=
\boldsymbol\omega_0\tau+
\frac{1}{2}\boldsymbol\alpha_\omega\tau^2,\\
\mathbf R(\tau)
&\simeq
\mathbf R_0
\operatorname{Exp}\!\left(
\Delta\boldsymbol\theta(\tau)^\wedge
\right),\\
\mathbf p(\tau)
&=
\mathbf p_0+\mathbf v_0\tau+
\frac{1}{2}\mathbf a_0\tau^2+
\frac{1}{6}\mathbf j\tau^3 .
\end{aligned}
\label{eq:continuous_reconstruction}
\end{equation}
Successive IMU intervals form the predicted trajectory
$\mathbf T_{OL}^{\rm pred}(t)$ over $[t_s,t_e]$.
The terminal posterior correction is
$\Delta\mathbf T_e=
\mathbf T_{OL}^{\rm post}(t_e)
(\mathbf T_{OL}^{\rm pred}(t_e))^{-1}$.
The frontend smoothly distributes this correction over the trajectory
using $\alpha=(t-t_s)/(t_e-t_s)$ and
$\psi(\alpha)=10\alpha^3-15\alpha^4+6\alpha^5$:
\begin{equation}
\mathbf T_{OL}(t)
=
\operatorname{Exp}\!\left[
\psi(\alpha)\operatorname{Log}(\Delta\mathbf T_e)
\right]
\mathbf T_{OL}^{\rm pred}(t).
\label{eq:posterior_reconstruction}
\end{equation}
At each complete physical revolution, a separate geometric observer
compares an independent IMU terminal prediction with the final LiDAR
posterior to apply bounded velocity and bias corrections without changing
pose. Accepted corrections affect subsequent propagation and trigger
motion compensation of the completed sweep before keyframe transfer.

\subsection{Keyframe Interface}
\label{sec:keyframe_interface}

The asynchronous keyframe interface selects completed sweeps and bounds
pending work so the slower neural backend does not block high-rate
local estimation.

The $k$-th full sweep spans $[t_k^s,t_k^e]$.
The frontend motion-compensates each LiDAR point $\mathbf p_c^L$ from
its acquisition time $t_c\in[t_k^s,t_k^e]$ into the sweep-end LiDAR frame:
\begin{equation}
\widetilde{\mathbf p}_c
=
\mathbf T_{OL}^{-1}(t_k^e)
\odot
\left(
\mathbf T_{OL}(t_c)\odot\mathbf p_c^L
\right).
\label{eq:final_sweep}
\end{equation}
Each candidate is
$(\mathcal P_k,[t_k^s,t_k^e],\mathbf T_k,\boldsymbol\Sigma_k)$,
where $\mathcal P_k$ contains the motion-compensated points,
$\mathbf T_k=\mathbf T_{OL}(t_k^e)$ is the terminal pose, and
$\boldsymbol\Sigma_k$ is the $6\times6$ attitude--position posterior
covariance.
After neural-map initialization, the interface allows at most one
keyframe in flight and one candidate pending. A candidate must satisfy
a minimum time gap and a translation, rotation, or elapsed-time condition.
While the backend is busy, newer eligible candidates replace the pending
one, and the interface discards stale candidates.

\subsection{Neural Implicit Backend}
\label{sec:neural_backend}

\subsubsection{Neural Implicit Map}

The backend represents the environment with a sparse point-based neural
implicit map $\mathcal N=\{\mathbf n_j\}_{j=1}^{N_n}$~\cite{pan2024pinslam},
where $N_n$ is the number of neural points and
$\mathbf n_j=(\boldsymbol\mu_j,\mathbf R_j,\mathbf h_j,c_j)$ stores its
position, local orientation, latent feature, and training support.
The first selected sweep $\mathcal P_k$ initializes the map in the
gravity-aligned frame $M$ with zero yaw and translation.
For a query $\mathbf q_i$ in $M$, spatial hashing retrieves up to $K$
neighboring neural points with index set $\mathcal K_i$. The backend
interpolates their features and decodes the SDF value as
\begin{equation}
\begin{aligned}
w_{ij}
&=
\frac{(\|\mathbf q_i-\boldsymbol\mu_j\|^2+\epsilon)^{-1}}
{\sum_{l\in\mathcal K_i}
(\|\mathbf q_i-\boldsymbol\mu_l\|^2+\epsilon)^{-1}},\\
\mathbf z_i
&=
\sum_{j\in\mathcal K_i}w_{ij}
\begin{bmatrix}
\mathbf h_j\\
\mathbf R_j^T(\mathbf q_i-\boldsymbol\mu_j)
\end{bmatrix},
\qquad
F(\mathbf q_i)=f_\theta(\mathbf z_i) ,
\end{aligned}
\label{eq:neural_field}
\end{equation}
where $\epsilon>0$ regularizes the weights, and the multilayer
perceptron (MLP) decoder
$f_\theta$ defines the neural SDF $F$.
The backend initializes the support score to $c_j=0$ and accumulates
it as $c_j\leftarrow c_j+\sum_i w_{ij}$. Mapping prioritizes
lower-support neural points.

\subsubsection{Neural Tracking}
Neural tracking aligns each selected keyframe with the current neural
map before mapping. The committed backend pose of keyframe $i$ is
$\mathbf T_i^B=\mathbf T_{ML,i}$. Since asynchronous transfer may skip
candidates, initialization uses the most recently committed keyframe
$p(k)$. Its frontend relative motion to keyframe $k$,
$\Delta\mathbf T_k^{\rm Front}=\mathbf T_{p(k)}^{-1}\mathbf T_k$,
initializes tracking as
$\mathbf T_k^{\rm init}=
\mathbf T_{p(k)}^B\Delta\mathbf T_k^{\rm Front}$.
To schedule tracking effort, the backend computes normalized uncertainty
from the rotational and translational covariance blocks
$\boldsymbol\Sigma_{R,k}$ and $\boldsymbol\Sigma_{p,k}$ of
$\boldsymbol\Sigma_k$:
$u_k=\max(\operatorname{tr}(\boldsymbol\Sigma_{R,k})/\tau_R,
\operatorname{tr}(\boldsymbol\Sigma_{p,k})/\tau_p)$.
The first convergence-check iteration is
\begin{equation}
n_{c,k}
=
\operatorname{clip}\!\left(
\left\lceil
n_{\min}[\max(1,\beta u_k)]^\gamma
\right\rceil,
n_{\min},n_{\max}
\right) ,
\label{eq:adaptive_tracking}
\end{equation}
where $\tau_R,\tau_p$ normalize uncertainty, $\beta,\gamma$ control
the schedule, and $n_{\min},n_{\max}$ bound it. Higher uncertainty
delays the first convergence check; the backend uses $n_{\max}$ when the covariance is invalid
Tracking continues until convergence or the iteration limit $n_{\max}$.
Covariance changes only the schedule, adding no pose-prior loss.

For $\mathbf p_i\in\mathcal P_k$ and current pose $\mathbf T$,
$\mathbf q_i=\mathbf T\odot\mathbf p_i$ gives the map-frame query,
with residual $r_i=F(\mathbf q_i)$ and gradient
$\mathbf g_i=\nabla_{\mathbf q_i}F(\mathbf q_i)$.
For the left-multiplicative perturbation
$\delta\boldsymbol\xi=
[\delta\boldsymbol\theta^T,\delta\mathbf t^T]^T$, the Jacobian is
$\mathbf J_i^{\rm N}=[(\mathbf q_i\times\mathbf g_i)^T\;\mathbf g_i^T]$.
Each valid query has a robust weight $\omega_i$ that downweights
large SDF residuals and gradient norms that deviate from one.
Stacking valid residuals and Jacobians into $\mathbf r$ and
$\mathbf J^{\rm N}$, with $\mathbf W=\operatorname{diag}(\omega_i)$, gives
\begin{equation}
\left[
(\mathbf J^{\rm N})^T\mathbf W\mathbf J^{\rm N}+
\lambda\operatorname{diag}((\mathbf J^{\rm N})^T\mathbf W\mathbf J^{\rm N})
\right]\delta\boldsymbol\xi
=
-(\mathbf J^{\rm N})^T\mathbf W\mathbf r ,
\label{eq:neural_tracking}
\end{equation}
where $\lambda>0$ is the damping coefficient. The backend updates the
pose as $\mathbf T\leftarrow\operatorname{Exp}(\delta\boldsymbol\xi)\mathbf T$.
If tracking fails, the backend skips map insertion; an early failure
also restores $\mathbf T_k^{\rm init}$.

\subsubsection{Neural Mapping}

After successful tracking, the backend samples near-surface and
free-space points along LiDAR rays. Their signed surface offsets $d_i$
are positive before the return and negative behind it. The targets
$y_i=\sigma(d_i/s)$ use the sigmoid function $\sigma$ and SDF transition
scale $s>0$. Mapping minimizes binary cross-entropy with logits
($\operatorname{BCEWithLogits}$) and Eikonal regularization:
\begin{equation}
\begin{aligned}
\mathcal L
&=
\frac{1}{N_s}
\sum_{i=1}^{N_s}
|\eta_i|\,
\operatorname{BCEWithLogits}
\left(
\frac{F(\mathbf q_i)}{s},y_i
\right)\\
&\quad+
\frac{\lambda_e}{N_e}
\sum_{i=1}^{N_e}
\left(
\|\nabla_{\mathbf q}F(\mathbf q_i)\|_2-1
\right)^2 ,
\end{aligned}
\label{eq:neural_mapping}
\end{equation}
where $N_s$ and $N_e$ denote the numbers of SDF and Eikonal samples,
$\eta_i$ is the sample weight, and $\lambda_e$ controls Eikonal
regularization. The first term fits the LiDAR-derived targets, while
the second encourages the SDF gradient to have unit norm.
Training combines current observations, replay samples, and samples
from regions with limited training support. After initial mapping,
the backend freezes the MLP decoder $f_\theta$ and continues to optimize
local neural-point features.
Fused CUDA operators combine queries, neighborhood aggregation, and
MLP decoding to reduce intermediate tensors, kernel launches, and GPU
memory traffic. Tracking computes gradients without building a general
automatic-differentiation graph. Dedicated forward and input-gradient
paths accelerate mapping with the frozen decoder while retaining
neural-point feature training. Preallocated workspaces reduce repeated
allocation and help stabilize latency.
These optimizations
preserve the tracking and mapping objectives.

\subsubsection{Pose Graph Optimization}

Pose graph optimization (PGO) corrects drift in $\{\mathbf T_i^B\}$
using sequential and accepted loop edges $(i,j)\in\mathcal E$.
Each edge has a relative-pose measurement $\mathbf Z_{ij}$ and
information matrix $\boldsymbol\Omega_{ij}$; sequential measurements
come from backend estimates. PGO minimizes
\begin{equation}
\{\mathbf T_i^{B,\star}\}
=
\arg\min_{\{\mathbf T_i^B\}}
\sum_{(i,j)\in\mathcal E}
\left\|
\operatorname{Log}\!\left[
\mathbf Z_{ij}^{-1}
(\mathbf T_i^B)^{-1}\mathbf T_j^B
\right]
\right\|_{\boldsymbol\Omega_{ij}}^2 ,
\label{eq:pgo}
\end{equation}
where
$\|\mathbf e\|_{\boldsymbol\Omega}^2
=\mathbf e^T\boldsymbol\Omega\mathbf e$.
Loop retrieval and neural-map deformation follow~\cite{pan2024pinslam}.
The backend applies accepted pose corrections to both the keyframe
trajectory and neural map, keeping them consistent within the same
global revision.

\subsubsection{Map-Odometry Correction}

For each accepted backend revision $\nu$, the backend and frontend poses
$\mathbf T_k^{B,(\nu)}$ and $\mathbf T_k$ of the same keyframe determine
the map-to-odometry transform. This transform aligns each new frontend
pose with the global map:
\begin{equation}
\begin{aligned}
\mathbf T_{MO}^{(\nu)}
&=
\mathbf T_k^{B,(\nu)}\mathbf T_k^{-1},\\
\mathbf T_{ML}^{\rm online}(t)
&=
\mathbf T_{MO}^{(\nu(t))}\mathbf T_{OL}(t),
\end{aligned}
\label{eq:global_correction}
\end{equation}
Here, $\nu(t)$ denotes the latest accepted backend revision at time $t$.
Between revisions, $\mathbf T_{MO}$ remains fixed. The system outputs
global poses at the frontend rate without changing frontend states
or historical IESKF estimates.

\section{EXPERIMENTS}
\label{sec:experiments}

\begin{table*}[t]
\caption{Trajectory accuracy as ATE RMSE
and mean profiling time per LiDAR frame.}
\label{tab:slam}
\centering
\footnotesize
\setlength{\tabcolsep}{3.5pt}
\renewcommand{\arraystretch}{1.1}
\begin{tabular*}{\textwidth}{@{}l@{\extracolsep{\fill}}cccccccc@{}}
\toprule
& \multicolumn{5}{c}{Newer College (Stereo-Cam) 64-channel}
& \multicolumn{3}{c}{M2DGR 32-channel} \\
\cmidrule(lr){2-6}\cmidrule(lr){7-9}
Method & NCD01 & NCD02 & NCD05 & NCD07 & \textbf{Mean}
& Hall05 & Door01 & \textbf{Mean} \\
\midrule
\multicolumn{9}{c}{\textbf{ATE RMSE [m] $\downarrow$}} \\
\midrule
PIN-SLAM
& 0.357 & 0.351 & 0.217 & 0.319 & 0.311
& \underline{1.031} & \textbf{0.271} & \underline{0.651} \\
CLID-SLAM
& 0.859 & 1.139 & \underline{0.126} & 0.231 & 0.589
& 1.182 & 0.433 & 0.808 \\
FAST-LIO2
& \underline{0.344} & 0.357 & \textbf{0.108}
& \underline{0.136} & \underline{0.236}
& 1.159 & 0.458 & 0.808 \\
Point-LIO
& 0.347 & \underline{0.317} & 0.192 & \textbf{0.135}
& 0.248 & 1.158 & 0.447 & 0.802 \\
\textbf{LIN-SLAM}
& \textbf{0.284} & \textbf{0.314} & 0.139 & 0.148
& \textbf{0.221}
& \textbf{1.005} & \underline{0.272} & \textbf{0.638} \\
\midrule
\multicolumn{9}{c}{\textbf{Mean profiling time [ms/frame] $\downarrow$}} \\
\midrule
PIN-SLAM
& 280.604 & 317.803 & \underline{184.472} & 246.553
& 257.358
& 233.624 & 267.413 & 250.518 \\
CLID-SLAM
& \underline{201.934} & \underline{281.923} & 270.491
& \underline{239.437} & \underline{248.446}
& \underline{193.991} & \underline{167.727} & \underline{180.859} \\
\textbf{LIN-SLAM}
& \textbf{42.034} & \textbf{54.130} & \textbf{52.119}
& \textbf{48.562} & \textbf{49.211}
& \textbf{62.279} & \textbf{58.559} & \textbf{60.419} \\
\bottomrule
\end{tabular*}

\par\smallskip
\begin{tabular*}{\textwidth}{@{}l@{\extracolsep{\fill}}ccccccc@{}}
\toprule
& \multicolumn{7}{c}{Newer College (Multi-Cam) 128-channel} \\
\cmidrule(lr){2-8}
Method
& math\_easy & math\_medium & math\_hard
& quad\_easy & quad\_medium & quad\_hard & \textbf{Mean} \\
\midrule
\multicolumn{8}{c}{\textbf{ATE RMSE [m] $\downarrow$}} \\
\midrule
PIN-SLAM
& \underline{0.091} & \textit{Fail} & \textit{Fail}
& 0.095 & \textit{Fail} & 0.176 & 0.121 \\
CLID-SLAM
& \textbf{0.071} & \textbf{0.115} & 0.090
& 0.078 & \underline{0.064} & 0.095 & \textbf{0.086} \\
FAST-LIO2
& 0.112 & \textbf{0.115} & \underline{0.084}
& \textbf{0.067} & \textbf{0.058} & 0.151 & 0.098 \\
Point-LIO
& 0.175 & 0.117 & 0.089
& \underline{0.071} & 0.079 & \textbf{0.060} & 0.099 \\
\textbf{LIN-SLAM}
& 0.096 & 0.141 & \textbf{0.074}
& 0.073 & 0.065 & \underline{0.069} & \textbf{0.086} \\
\midrule
\multicolumn{8}{c}{\textbf{Mean profiling time [ms/frame] $\downarrow$}} \\
\midrule
PIN-SLAM
& 175.782 & \textit{Fail} & \textit{Fail}
& \underline{181.177} & \textit{Fail} & 196.198
& \underline{184.386} \\
CLID-SLAM
& \underline{173.225} & \underline{203.845}
& \underline{191.416} & 193.267
& \underline{197.585} & \underline{188.753} & 191.349 \\
\textbf{LIN-SLAM}
& \textbf{55.383} & \textbf{64.937}
& \textbf{64.084} & \textbf{57.688}
& \textbf{62.385} & \textbf{66.095} & \textbf{61.762} \\
\bottomrule
\end{tabular*}

\par\smallskip
\begin{minipage}{\textwidth}
\scriptsize
One frame denotes one complete physical LiDAR revolution.
FAST-LIO2 processes complete LiDAR revolutions, whereas Point-LIO
performs point-wise updates.
Geometric baselines use latency-equivalent LiDAR data availability
for fair comparison.
\textbf{Bold}/\underline{underline}: best/second-best among reported
successful runs.
Means are arithmetic averages of per-sequence results within each
dataset group, excluding \textit{Fail} entries.
\end{minipage}
\end{table*}

\begin{table}[t]
\caption{3D reconstruction accuracy. Bold indicates the best result per
scene.}
\label{tab:reconstruction}
\centering
\footnotesize
\setlength{\tabcolsep}{3pt}
\renewcommand{\arraystretch}{1.1}
\begin{tabular}{lcccc}
\toprule
Method
& \shortstack{Accuracy\\{[cm]} $\downarrow$}
& \shortstack{Completeness\\{[cm]} $\downarrow$}
& \shortstack{Chamfer-L1\\{[cm]} $\downarrow$}
& \shortstack{F-score\\{[\%]} $\uparrow$} \\
\midrule
\multicolumn{5}{c}{Quad} \\
\midrule
PIN-SLAM  & 10.768 & 20.346 & 15.557 & 79.08 \\
CLID-SLAM & 12.060 & 15.435 & 13.748 & 86.06 \\
LIN-SLAM  & \textbf{8.940} & \textbf{13.684}
& \textbf{11.312} & \textbf{87.84} \\
\midrule
\multicolumn{5}{c}{Math Institute} \\
\midrule
PIN-SLAM  & 16.659 & 17.126 & 16.892 & 81.66 \\
CLID-SLAM & 15.833 & 16.163 & 15.998 & 82.86 \\
LIN-SLAM  & \textbf{14.523} & \textbf{15.455}
& \textbf{14.989} & \textbf{85.31} \\
\bottomrule
\end{tabular}
\end{table}

\begin{table}[t]
\caption{Controlled neural-backend execution comparison. Timing entries
report mean/95th-percentile (P95) latency.}
\label{tab:backend}
\centering
\footnotesize
\setlength{\tabcolsep}{5pt}
\renewcommand{\arraystretch}{1.05}
\begin{tabular}{lcc}
\toprule
Metric & Tensor implementation & Optimized  C++/CUDA \\
\midrule
Tracking [ms]
& 79.674/167.922 & \textbf{35.076/58.016} \\
Mapping [ms]
& 130.901/139.378 & \textbf{29.545/31.890} \\
Backend total [ms]
& 246.278/361.939 & \textbf{82.605/107.777} \\
Final ATE RMSE [m]
& 0.2918 & \textbf{0.2835} \\
\bottomrule
\end{tabular}
\end{table}

Public-dataset experiments compare localization accuracy, reconstruction
quality, and computational efficiency. Onboard experiments assess
real-time operation, resource utilization, and long-duration deployment.

\subsection{Public-Dataset Experiments}
\label{sec:accuracy_efficiency}

\subsubsection{Experimental Setup}

We evaluate twelve indoor and outdoor sequences: four from 64-channel
Newer College Stereo-Cam~\cite{ncd}, six from 128-channel Newer College
Multi-Cam~\cite{zhang2021multicamera}, and two from 32-channel
M2DGR~\cite{yin2022m2dgr} (Table~\ref{tab:slam}). NCD01 also supports
the controlled comparison of neural-backend execution.
The 128-channel Multi-Cam experiments adjust input downsampling,
tracking voxel size, processing range, and surface sampling.
All public-dataset experiments use dual Intel Xeon Gold 6530 CPUs and
an NVIDIA GeForce RTX 5090 GPU.

Localization accuracy is evaluated by the root mean square error (RMSE) of ATE~\cite{sturm2012benchmark} after Horn
$\mathrm{SE}(3)$ alignment~\cite{horn1987closed}. For estimated and
ground-truth positions $\hat{\mathbf p}_i$ and $\mathbf p_i$,
$\mathrm{ATE}=\sqrt{N^{-1}\sum_i
\|\mathbf p_i-(\mathbf R\hat{\mathbf p}_i+\mathbf t)\|_2^2}$.
Reconstruction quality is evaluated using accuracy, completeness,
Chamfer-L1, and F-score~\cite{pan2024pinslam}. Accuracy and completeness measure the two
directed nearest-neighbor distances between the reconstructed and
reference surfaces, and Chamfer-L1 is their average. F-score is the
harmonic mean of precision and recall at a fixed distance threshold.
PIN-SLAM, CLID-SLAM, and LIN-SLAM use the same reconstruction resolution,
cropping region, surface sampling, and F-score distance threshold.

The baselines include neural implicit methods
PIN-SLAM~\cite{pan2024pinslam} and CLID-SLAM~\cite{jiang2025clidslam},
and geometric LiDAR--inertial methods FAST-LIO2~\cite{xu2022fastlio2}
and Point-LIO~\cite{he2023pointlio}.

\subsubsection{Localization Accuracy and Runtime}

Table~\ref{tab:slam} shows that LIN-SLAM achieves the lowest mean ATE
RMSE on Newer College Stereo-Cam, Newer College Multi-Cam, and M2DGR.
On Stereo-Cam, the mean profiling time per frame decreases from
257.358~ms for PIN-SLAM and 248.446~ms for CLID-SLAM to 49.211~ms for
LIN-SLAM, corresponding to reductions of 80.9\% and 80.2\%, respectively.

Hall05's advantage without loop closures and the lowest mean
errors on the 32- and 64-channel datasets are consistent with improved
backend inputs from the geometric frontend and subsequent neural
tracking. Selective full-sweep admission reduces backend update
frequency; downsampling, processing range, sampling, and keyframe
admission explain why workload does not scale with channel count alone.

\begin{figure*}[t]
\centering
\includegraphics[width=\textwidth]{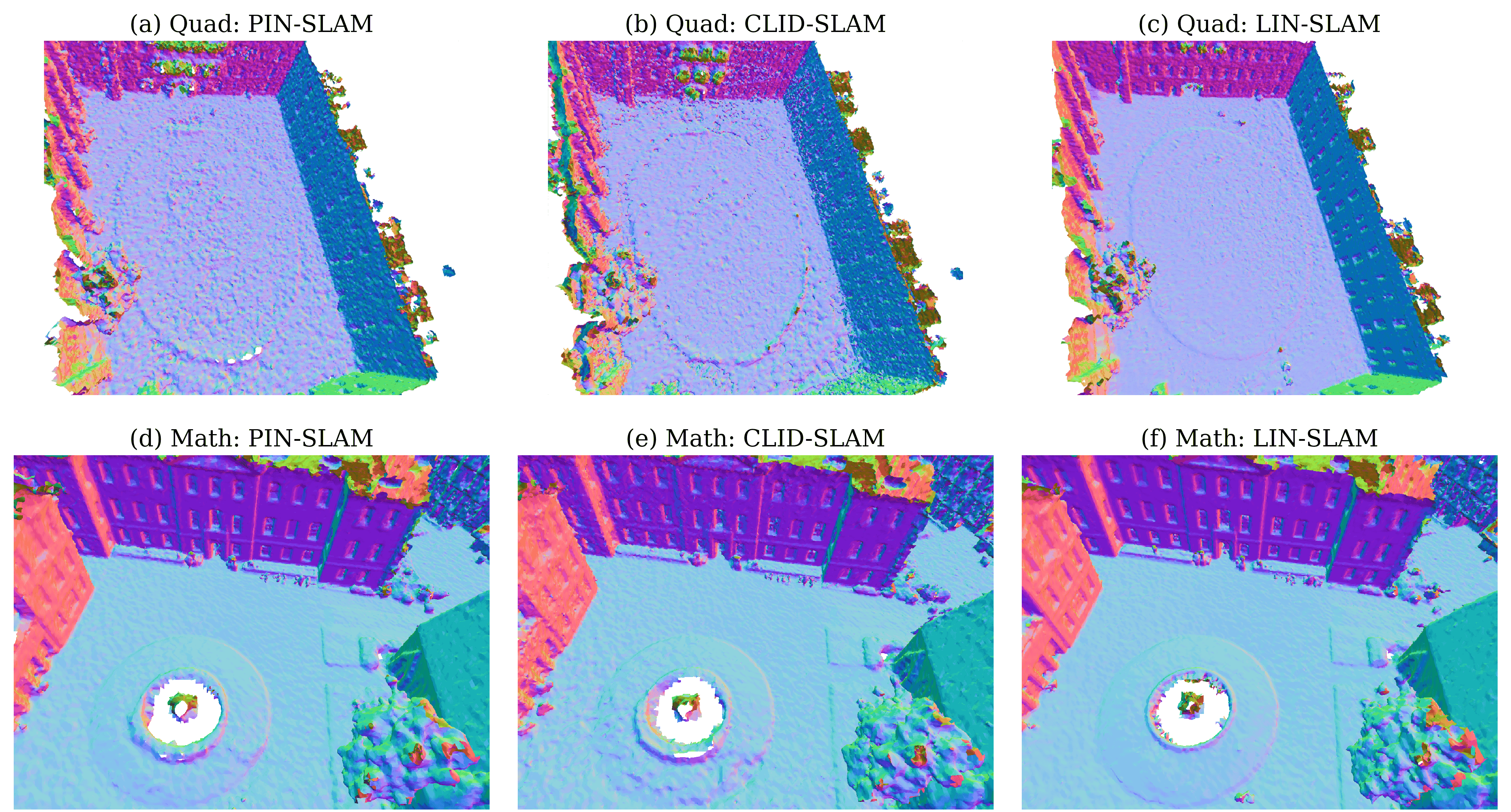}
\caption{Neural-SDF surface comparison on 64-channel Quad (top row) and
128-channel Math Institute (bottom row), using the same viewpoint within each scene.
PIN-SLAM: (a), (d); CLID-SLAM: (b), (e);
LIN-SLAM: (c), (f).}
\label{fig:ncd01_surface_comparison}
\end{figure*}

\subsubsection{Reconstruction Quality}

Table~\ref{tab:reconstruction} and
Fig.~\ref{fig:ncd01_surface_comparison} evaluate neural reconstruction
quality on Quad and Math Institute. LIN-SLAM achieves the lowest
accuracy, completeness, and Chamfer-L1 errors and the highest F-score
in both scenes.
In Fig.~\ref{fig:ncd01_surface_comparison}(c), LIN-SLAM reduces ground
ghosting and local ripples in Quad, with clearer edges and more
continuous connections to surrounding structures. Math Institute shows
smaller visual differences, mainly in ground uniformity and local
boundaries in Fig.~\ref{fig:ncd01_surface_comparison}(f).

The larger improvement on 64-channel Quad is consistent with better
alignment from frontend motion compensation and relative-motion
initialization under sparser observations, although scene and
configuration differences prevent attributing it to channel count alone.

\subsubsection{Neural-Backend Execution Efficiency}

To evaluate execution speed and localization accuracy together, we
compare tensor and optimized C++/CUDA implementations on NCD01 using
identical keyframes, initialization, samples, network architecture,
replay data, and optimization settings.

Table~\ref{tab:backend} shows substantially faster execution without
loss of localization accuracy in this comparison. Mean backend latency
falls from 246.278 to 82.605~ms (66.5\%), and P95
latency falls from 361.939 to 107.777~ms (70.2\%), while final ATE RMSE
remains comparable at 0.2918 and 0.2835~m, respectively.
Mapping shows the largest reduction, from 130.901 to 29.545~ms.

Fewer intermediate tensors, kernel launches, and allocations, together
with lower tracking-gradient overhead, reduce execution cost; dedicated forward and input-gradient paths
accelerate mapping with the frozen decoder while retaining neural-point
feature training and the original objectives.

\subsection{Onboard Deployment Experiments}
\label{sec:onboard}

\begin{figure}[t]
\centering
\includegraphics[width=\columnwidth]{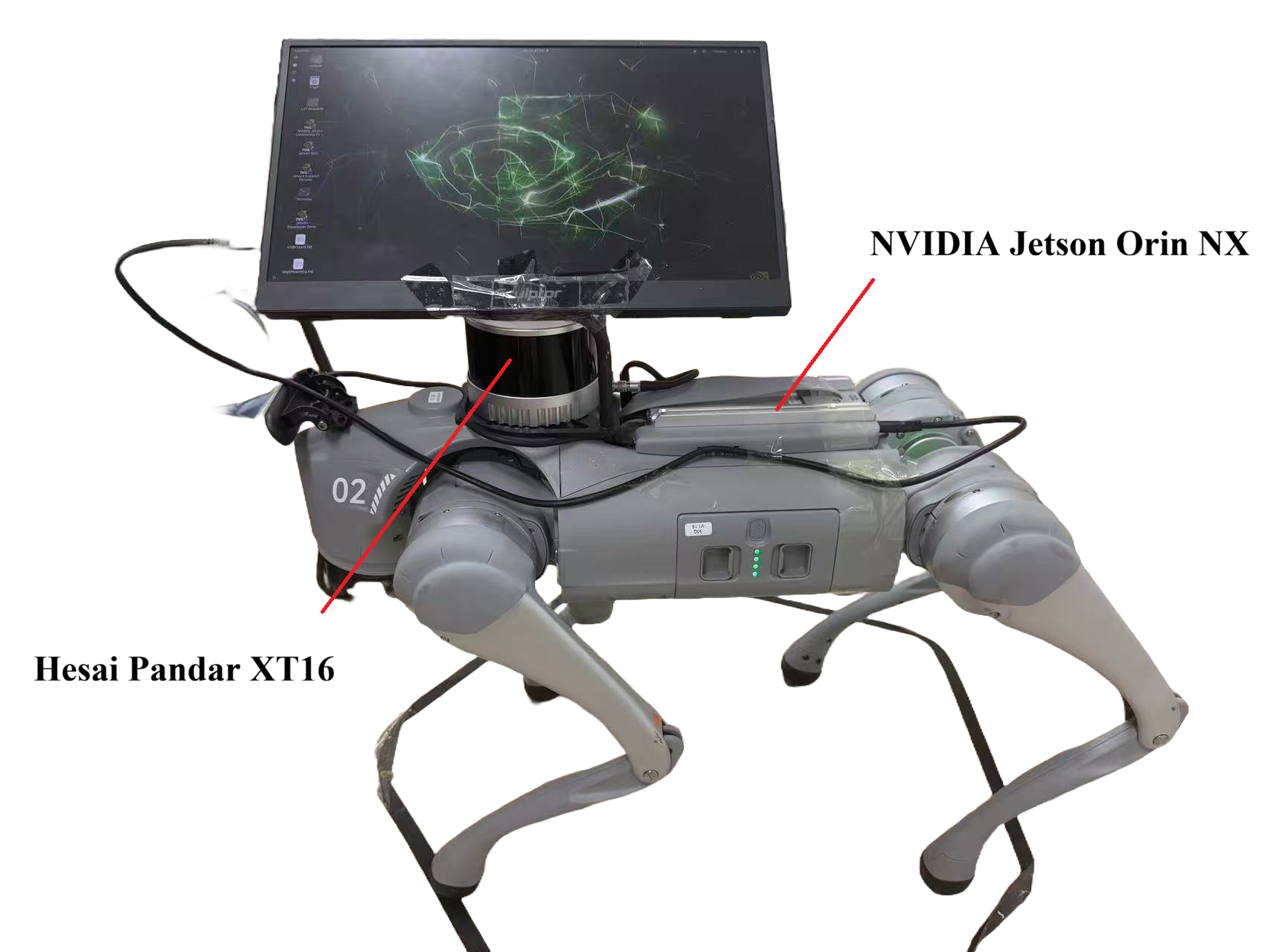}
\caption{Unitree Go2 platform used for the onboard experiment.}
\label{fig:go2_platform}
\end{figure}

\subsubsection{Experimental Setup}

The Unitree Go2 platform (Fig.~\ref{fig:go2_platform}) runs the complete
LIN-SLAM system with a 16-channel Hesai Pandar XT16 LiDAR and an NVIDIA
Jetson Orin NX. Corridor covers approximately 1.47~km in 22.03~min
within a $216\times156$~m area with long corridors, repetitive layouts,
and multiple revisits. Basement covers approximately 0.86~km in
13.99~min within a $92\times108$~m area with repetitive parking aisles,
pillars, vehicles, and partial occlusions.
LIN-SLAM runs the complete pipeline concurrently at the original
$1\times$ sensor rate. PIN-SLAM and CLID-SLAM cannot sustain this rate
with localization and continual neural mapping under the tested Orin NX
configuration, so the comparison uses FAST-LIO2 and Point-LIO as
geometric baselines.

Without external ground truth, evaluation uses endpoint closure error
and qualitative map structure. Endpoint closure error measures the
Euclidean distance between estimated start and end positions after
returning to the start, indicating endpoint consistency rather than
trajectory-wide accuracy. LIN-SLAM uses its final globally corrected
trajectory after loop closure; the geometric baselines use their final
odometry trajectories. Profiling time per frame equals total profiled 
processing time divided by the number of complete physical LiDAR revolutions.
We report mean/peak CPU and GPU utilization over each sequence.

\begin{table*}[t]
\caption{Onboard deployment comparison on the Go2/Orin NX platform.
Resource-utilization entries report mean/peak values.}
\label{tab:onboard}
\centering
\scriptsize
\setlength{\tabcolsep}{2.5pt}
\renewcommand{\arraystretch}{1.08}
\begin{tabular*}{\textwidth}{@{}l@{\extracolsep{\fill}}cccccc@{}}
\toprule
& \multicolumn{3}{c}{Corridor (1.47~km)}
& \multicolumn{3}{c}{Basement (0.86~km)} \\
\cmidrule(lr){2-4}\cmidrule(lr){5-7}
Metric
& LIN-SLAM & FAST-LIO2 & Point-LIO
& LIN-SLAM & FAST-LIO2 & Point-LIO \\
\midrule
\textbf{Endpoint closure error [m] $\downarrow$}
& \textbf{0.141} & \textit{Fail} & 34.144
& \textbf{0.073} & \textit{Fail} & 0.614 \\
GPU utilization mean/peak [\%]
& 82.4/99 & 0/0 & 0/0
& 86.9/99 & 0/0 & 0/0 \\
CPU utilization mean/peak [\%]
& 24.42/71.88 & 12.23/100.00 & 9.68/100.00
& 27.46/100.00 & 8.22/100.00 & 9.67/100.00 \\
Profiling time per frame [ms]
& 84.875 & 59.900 & 4.727
& 88.401 & 30.688 & 7.924 \\
\bottomrule
\end{tabular*}

\par\smallskip
\begin{minipage}{\textwidth}
\scriptsize
One frame denotes one complete physical LiDAR revolution.
\textit{Fail} denotes severe trajectory divergence that prevents a valid
endpoint closure measurement.
\end{minipage}
\end{table*}

\begin{figure*}[t]
\centering
\includegraphics[width=\textwidth]{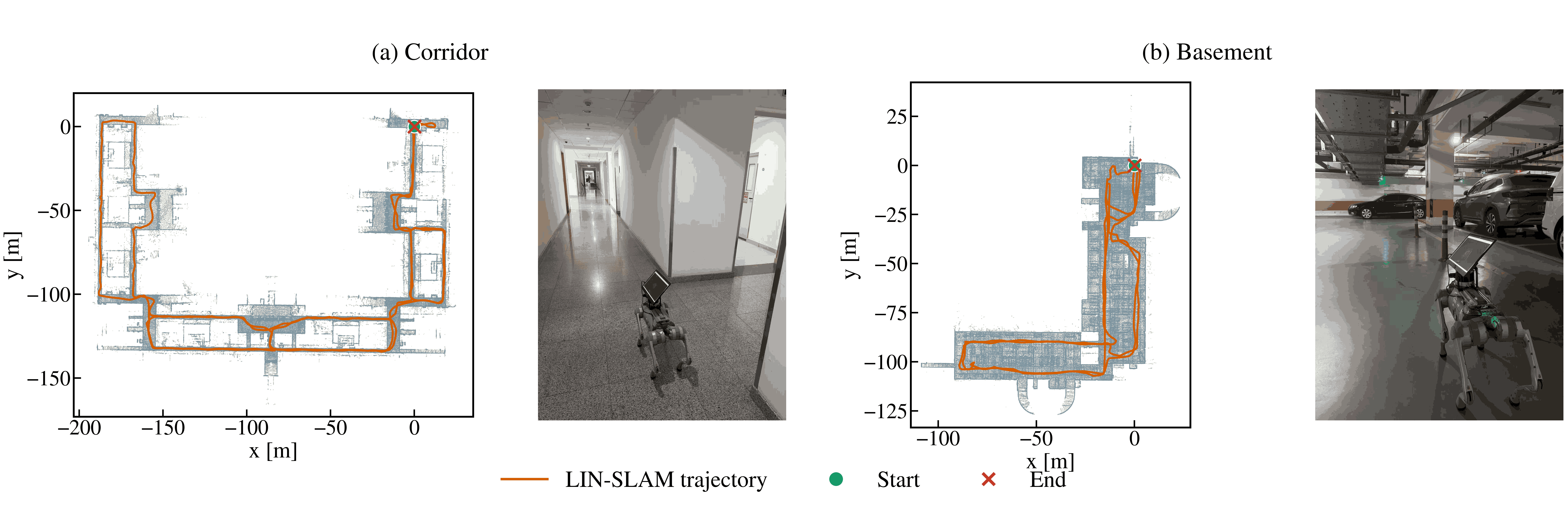}
\caption{Neural point maps produced during the onboard deployments:
(a) Corridor and (b) Basement.}
\label{fig:go2_onboard}
\end{figure*}

\subsubsection{Deployment Results}

LIN-SLAM operates stably throughout both sequences. As shown in
Table~\ref{tab:onboard}, its endpoint closure errors are 0.141 and
0.073~m, respectively. Mean profiling times of 84.875 and 88.401~ms
remain below the 100-ms period of the 10-Hz LiDAR stream, with mean
GPU utilization of 82.4\% and 86.9\%. FAST-LIO2 diverges early in
both sequences, whereas Point-LIO accumulates endpoint closure errors
of 34.144 and 0.614~m. The runs include 23 accepted loop closures in
Corridor and 22 in Basement.
Figure~\ref{fig:go2_onboard} shows aligned walls and continuous corridor
connections in Corridor, and pillar rows without visible ghosting in
Basement.

Fewer glass surfaces and weaker wall reflections may explain Basement's
better reconstruction. The geometric baselines' divergence and drift
show that geometric methods alone cannot preserve long-term consistency
in these runs, supporting the value of combination of high-rate local estimation,
neural tracking, and global map refinement.

\section{CONCLUSION}

We present LIN-SLAM, a lightweight LiDAR--inertial neural implicit SLAM system
that combines geometric estimation, asynchronous coupling, and
optimized C++/CUDA neural execution to enable real-time localization and continual
neural mapping.
We show that our method achieves state-of-the-art LiDAR-IMU localization accuracy on both datasets and onboard runs, while supporting online deployment on an NVIDIA Jetson Orin NX.
Future work will focus on combining LIN-SLAM with smooth control.

\end{document}